\documentclass[letterpaper, 10 pt, conference]{ieeeconf}

\IEEEoverridecommandlockouts                              

\usepackage{caption}
\usepackage{graphicx}        
\usepackage{overpic}
\usepackage{rotating}
\usepackage{amssymb}
\usepackage{amsfonts}
\usepackage{amsmath} 
\usepackage[usenames,dvipsnames]{color}
\usepackage{verbatim}
\usepackage[ruled]{algorithm}
\usepackage[noend]{algorithmic}

\usepackage{morefloats}        
\usepackage{xspace}            

\usepackage[dvips,frame,rotate,import]{xy}

\usepackage{hyperref} 

\usepackage{xr}
\graphicspath{{./}}

\newcommand{\lcma}{{\,,}}
\newcommand{\rcma}{{,\,}}

\newcommand{\cma}{{\lcma}}
\newcommand{\lprd}{{\,.}}

\newcommand{\prd}{{\lprd}}

\providecommand{\abs}[1]{{| #1 |}}

\providecommand{\norm}[1]{{\| #1 \|}}

\providecommand{\pair}[2]{{#1 \cdot #2}}

\renewcommand{\Re}{\mathbb{R}}

\newcommand{\Xobs}{X_{\rm obs}}
\newcommand{\Xfree}{X_{\rm free}}

\newcommand{\Xgoal}{X_{\rm goal}}

\newcommand{\histu}{\tilde{u}} 
\newcommand{\histx}{\tilde{x}} 
\newcommand{\tf}{t_{\rm f}} 

\newcommand{\dif}[1]{\, \mathrm{d} #1 \,}

\newcommand{\Error}{{\mathcal{E}}}

\providecommand{\set}[1]{\{#1\}}

\providecommand{\cset}[2]{{\{{#1}\nobreak\mid\nobreak{#2}\}}}

\providecommand{\Bigcset}[2]{{\Big\{{#1}\nobreak\;\Big|\;\nobreak{#2}\Big\}}}

\providecommand{\intcc}[2]{[#1 \rcma #2]}

\newcommand{\simcom}{\mathcal{T}}

\newcommand{\simplex}{\tau}

\DeclareMathOperator{\St}{St}
\newcommand{\Xd}{\hat X}

\DeclareMathOperator*{\argmin}{arg\,min}

\DeclareMathOperator*{\Exp}{E}

\DeclareMathOperator{\minloc}{\mathbf{minloc}}

\providecommand{\ceil}[1]{\lceil {#1} \rceil}

\newcommand{\xRand}{x_{\rm new}}

\newcommand{\cost}{\hat V}
\newcommand{\costOf}[1]{\cost(#1)}

\newcommand{\lookAheadCost}{\cost^*}
\newcommand{\lookAheadCostOf}[1]{\cost^*(#1)}
\newcommand{\key}{K}

\newcommand{\RRGstar}{{RRG$^*$}\xspace}
\newcommand{\RRTstar}{{RRT$^*$}\xspace}
\newcommand{\RRTsharp}{{RRT$^\#$}\xspace}
\newcommand{\RRTx}{{RRT$^\mathrm{X}$}\xspace}
\newcommand{\PRMstar}{{PRM$^*$}\xspace}
\newcommand{\Astar}{{A$^*$}\xspace}
\newcommand{\Dstar}{{D$^*$}\xspace}

\newcommand{\PDE}{{partial differential equation}\xspace}

\newcommand{\HJB}{{Hamilton-Jacobi-Bellman}\xspace}

\newcommand{\Lebeg}{{Lebesgue}\xspace}

\providecommand{\defref}[1]{Definition~\ref{#1}}
\providecommand{\lemref}[1]{Lemma~\ref{#1}}

\providecommand{\secref}[1]{Section~\ref{#1}}

\providecommand{\figref}[1]{Fig.~\ref{#1}}

\newcommand{\noHyperLink}[1]{\begin{NoHyper}#1\end{NoHyper}}

\providecommand{\algref}[1]{\noHyperLink{Algorithm~\ref{#1}}}
\providecommand{\linref}[1]{\noHyperLink{Line~\ref{#1}}}

\newcommand{\topic}[1]{}
\newcommand{\tocite}[1]{\cite{#1}}
\newcommand{\todo}[1]{}
\newcommand{\authorcomment}[2]{}
\newcommand{\rewriteline}[1]{}
\newcommand{\rewrite}[1]{}
\newcommand{\removeline}[1]{}
\newcommand{\remove}[1]{}
\newcommand{\difference}[2]{}

\newcommand{\Known}[1]{}
\newcommand{\New}[1]{}
\newcommand{\Old}[1]{}

\newcommand{\wcone}[1]{}
\newcommand{\wctwo}[1]{}
\newcommand{\wcthree}[1]{}
\newcommand{\wcfour}[1]{}
\newcommand{\wcfive}[1]{}
\newcommand{\wcsix}[1]{}
\newcommand{\wcseven}[1]{}
\newcommand{\wceight}[1]{}
\newcommand{\wcnine}[1]{}

\usepackage{fullpage}

\usepackage[top=0.75in,bottom=1.25in,hmargin=0.5in]{geometry}

\newtheorem{thm}{Theorem}

\newtheorem{theorem}[thm]{Theorem}

\newtheorem{lemma}[thm]{Lemma}
\newtheorem{definition}[thm]{Definition}

\title{Planning for Optimal Feedback Control in the Volume of Free Space}

\author{Dmitry Yershov \and Michael Otte \and Emilio Frazzoli
  \thanks{This work was supported by AFOSR FA8650-07-2-3744, ONR MURI
    N00014-09-1-1051, and ARO MURI W911NF-11-1-0046.}  
  \thanks{The Authors are with the Laboratory for Information and
    Decision Systems, Massachusetts Institute of Technology,
    Cambridge, MA 02139, USA {\tt\small
      [yershov,ottemw,frazzoli]@mit.edu}}}

\begin{document}

\maketitle

\thispagestyle{empty}
\pagestyle{empty}

\begin{abstract}
  The problem of optimal feedback planning among obstacles in
  $d$-dimensional configuration spaces is considered.
  We present a sampling-based, asymptotically optimal feedback
  planning method. Our method combines an incremental construction of
  the Delaunay triangulation, volumetric collision-detection module,
  and a modified Fast Marching Method to compute a converging sequence
  of feedback functions.
  The convergence and asymptotic runtime are proven theoretically and
  investigated during numerical experiments, in which the proposed
  method is compared with the state-of-the-art asymptotically optimal
  path planners. The results show that our method is competitive with
  the previous algorithms.
  Unlike the shortest trajectory computed by many path planning
  algorithms, the resulting feedback functions can be used directly
  for robot navigation in our case.
  Finally, we present a straightforward extension of our method that
  handles dynamic environments where obstacles can appear, disappear,
  or move. 

\end{abstract}

%
%

\section{INTRODUCTION}\label{sec:introduction}

\topic{ACIDIC method}

\begin{figure}[t]

  \footnotesize
  \hspace{1.1cm} 0.7 s \hspace{2.1cm} 1.5 s \hspace{2.1cm} 5 s

  \centering

  \includegraphics[height=3.2cm, trim=150 230 180 235, clip=true]{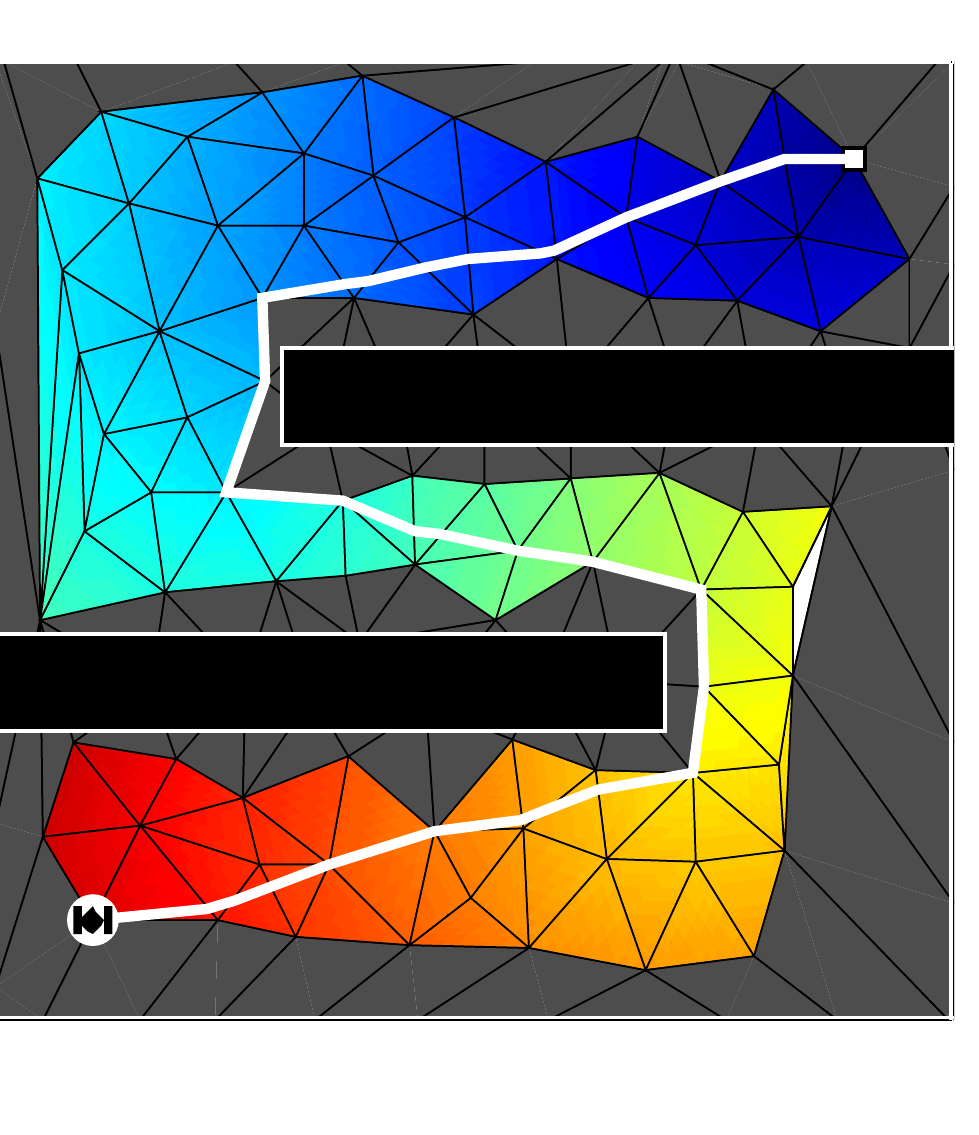}
  \includegraphics[height=3.2cm, trim=150 230 180 235, clip=true]{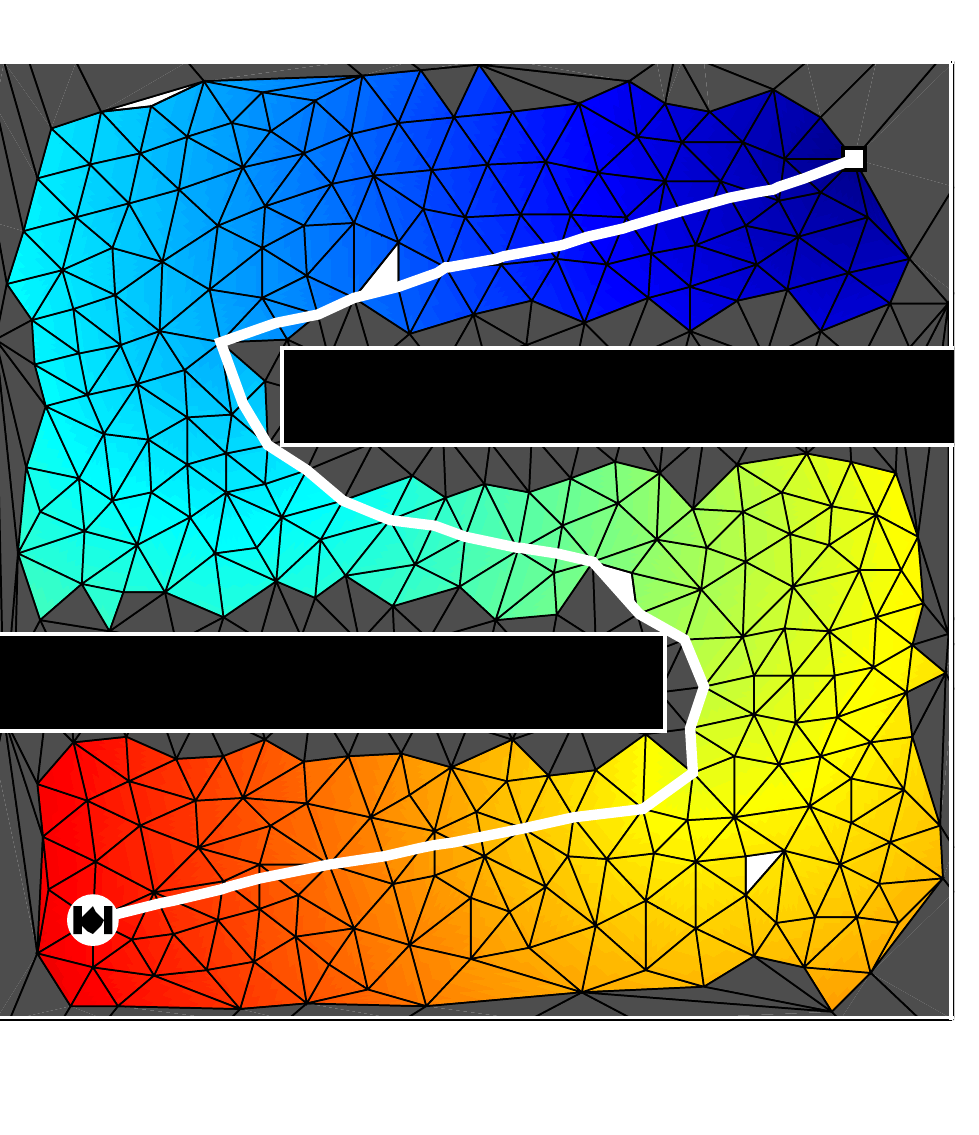}
  \includegraphics[height=3.2cm, trim=150 230 180 235, clip=true]{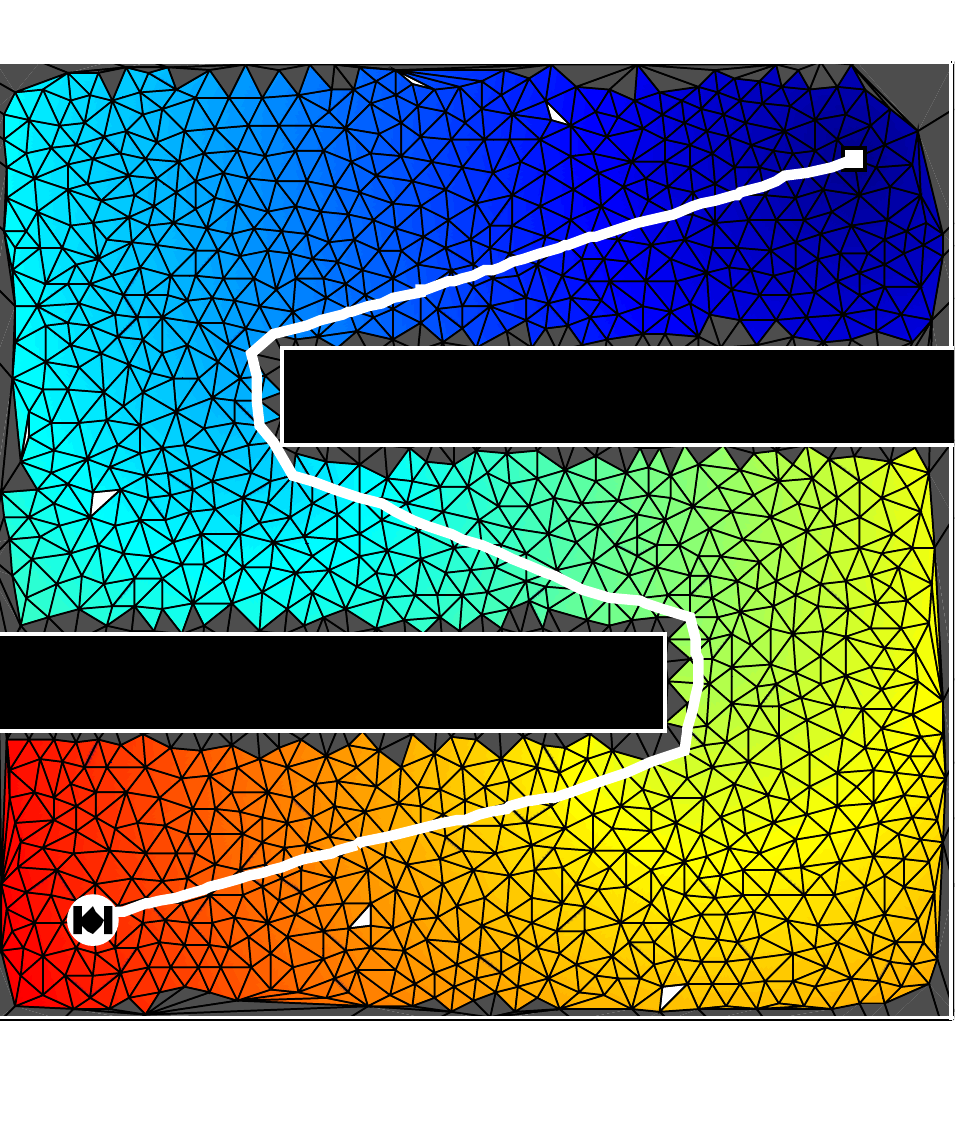}

  \caption[]{Incremental feedback plan (color) and resulting path
    (white) created by the proposed method for a robot (lower left)
    that desires to reach a goal (upper right) while avoiding
    obstacles (black). Regions for which a policy is still unknown are
    gray. Movies, including replanning in dynamic environments, appear
    at \url{http://tinyurl.com/qjnazvr}}

  \label{fig:abstract}
\end{figure}

We present the Asymptotically-optimal Control over Incremental
Delaunay sImplicial Complexes (ACIDIC) method for computing an optimal
feedback control among obstacles in $d$-dimensional configuration
spaces. In this method, a sample sequence, the Delaunay triangulation,
and a volumetric collision-detection module are used to build a
simplicial free space approximation. This volumetric approximation
parallels the Probabilistic RoadMap (PRM)~\tocite{KSLO96} in that it
captures the topology of the free space and has the same asymptotic
computational complexity as its optimal implementation
\PRMstar~\tocite{KarFra11}. Contrary to the PRM and \PRMstar, using
the Fast Marching Method (FMM)~\tocite{Sethian99} on this simplicial
approximation enables planning for a near-optimal path through the
volume of $d$-dimensional cells instead of constraining it to edges of
an 1D graph; see \figref{fig:abstract}. Moreover, borrowing ideas from
graph search literature~\tocite{KoeLik02, KoeLikFur04, Sten95}, we
extend the FMM to handle vertex insertions and deletions, which led us
to efficient asymptotically optimal feedback planning and replanning
algorithms.

\topic{Optimal planning in general}

For almost three 
decades, computing the optimal robot motion has been a central problem
in robotics. Optimal 
solutions are required when resources such as fuel and time are
expensive 
or limited. However, the optimal planning problem is computationally
hard.
For example, a relatively ``easy'' problem of finding the shortest
path in a $3$-dimensional polygonal environment is already
PSPACE-hard~\tocite{CaRe87}. Thus, only approximate optimal paths are
computed in practice.

\topic{Approximately optimal and Asymptotic optimality}

Recently, there has been a flurry of interest in sampling-based
methods that provide \emph{asymptotic optimality} by improving
iteratively near-optimal solutions.
For example, \RRGstar~\tocite{KarFra11} 
constructs a rapidly-exploring random graph (RRG) using a sample
sequence of random vertices.
%
%
In practice, such algorithms can be terminated after a sufficiently
accurate solution has been found. In this respect, they are similar to
anytime graph-search algorithms~\tocite{KoeLikFur04}. 
Contrary to graph-search algorithms, the planners such as RRG converge
to the true optimal path in the limit, as the iteration number tends
to infinity.

\topic{existing algorithms: path-centric vs. control-centric}

Most sampling-based planning algorithms are \emph{path-centric}, that
is, they find an
%
open-loop control that defines a near-optimal trajectory
from start to goal~\tocite{ArsTsi13, DobBer14, OttFra14,
  PPKKLP12}. When used for robot navigation, such algorithms require
an additional path-following controller to 
close the control loop. On the other hand, \emph{control-centric}
algorithms compute a feedback function that stabilizes the dynamical
system towards the goal---eliminating the need of auxiliary
controllers. Nonoptimal feedback planning algorithms are well
established in robotics~\tocite{LiLa09, RiKo92,TeMaToRo10}. The first
incremental, sampling-based, asymptotically optimal feedback
stochastic control planner is presented
in~\tocite{HuyKarFra12}. Introduced in~\tocite{YerLav12}, Simplicial
Dijkstra and \Astar algorithms compute approximate optimal feedback
control using a simplicial free space decomposition and the FMM, which
recently have been combined with the adaptive mesh refinement yielding
an asymptotically optimal, but not sampling-based feedback planning
algorithm~\tocite{YerFra14}.

%



\topic{we are the first FMM people and bash FMT*}

Our algorithm is different from the previous attempts to combine
sampling strategies with the FMM.
In~\cite{GoAlGaMo14}, for example, the RRT algorithm is used to
compute the initial path, which is improved during the post-processing
using the FMM over a local Delaunay triangulation.
Since only a local approximation is used, the convergence to the
globally optimal path is not guaranteed.
Despite a naming similarity, the Fast Marching Trees (FMT$^*$)
algorithm~\tocite{JanPav13} does not implement the FMM. The FMT$^*$
computes cost-to-go wavefront propagation using its values from the
neighboring vertices instead of interpolating between them.
Thus, it is closer in spirit to Dijkstra's graph search algorithm.
%

To the best of our knowledge, the ACIDIC is the first asymptotically
optimal feedback planning method that combines sampling strategies,
volumetric free space approximations, and efficient FMM-based feedback
planning algorithms.

\section{PRELIMINARIES}
\label{sec:prelim}

\subsection{Optimal Feedback Planning Problem} 
\label{sec:problem_formulation}

\topic{The stage and the actor}

An optimal control formulation 
provides the most natural framework for the feedback planning
algorithm. 
Let $X$ be the configuration space of a
robot, 
that is, the complete set of parameters and their respective ranges
that uniquely determine robot's internal configuration and position in
the world. The open \emph{obstacle set}, $\Xobs \subset X$,
corresponds to all states of the robot that result in either
self-collisions or collisions with obstacles in the world. The robot
is free to traverse the \emph{free space}, $\Xfree = X \setminus
\Xobs$, along trajectories of an ordinary differential equation (ODE)
with control
\begin{equation}
  \label{eq:system}
  \dot x(t) = f(x(t), u(t)) \prd
\end{equation}
This equation defines the dynamics of the robot and allows the
\emph{input signal} $u(t)$, chosen from the \emph{input set} $U$, to
control it. Finally, we assume that $\Xfree$ is a compact Lipschitz
domain, that is, the boundary of $\Xfree$ can be represented locally
as a level-set of a Lipschitz continuous function.

\topic{implicit definition of the free space using collision-detection
  module}

In many problems in robotics the configuration space $X$ and the
motion model \eqref{eq:system} are well-defined due to careful
mechanism design. The obstacle set, on the other hand, inherits the
complexity of the surrounding world, which is less predictable and may
also evolve over time. 



\topic{The optimality}

The performance of the robot is measured with respect to an additive
cost functional:
\begin{equation}
  \label{eq:cost_functional}
  J(\histx,\histu) = \int^{\tf}_{0} c(\histx(t), \histu(t)) \dif t 
\end{equation}
if $\histx(t) \in \Xfree$ for all $t \in \intcc{0}{\tf}$, and
$J(\histx,\histu) = \infty$ otherwise. In the above, $\histu$ and
$\histx$ are realizations of the input signal and the corresponding
trajectory, $c:X \times U \to \Re^+$ is a positive local cost, and
$\tf$ is the terminal time. Usually, $\tf$ is the first moment when
$\histx(\tf)$ is inside the \emph{goal set} $\Xgoal$. We assume,
$\Xgoal$ is a compact Lipschitz domain in $\Xfree$. The task of
optimal motion planning algorithm is to minimize $J$.

\topic{The cost-to-go and feedback control}

We employ Bellman's dynamic programming principle and find the optimal
control using the \emph{cost-to-go} function $V$. At all points $x \in
X$, $V(x)$ is defined as the optimal cost of reaching the goal from
$x$, and it satisfies \HJB ~\PDE (HJB PDE):
\begin{equation}
  \label{eq:hjb}
  \min_{u \in U} \set{\pair{\nabla V(x)}{f(x,u)} + c(x,u)} = 0 \prd
\end{equation}
Once the cost-to-go is computed, the feedback control is given as a
minimizing argument in \eqref{eq:hjb}, $\pi(x) = \argmin_{u \in U}
\set{\pair{\nabla V(x)}{f(x,u)} + c(x,u)}$. The optimal trajectory is
then found by replacing $u$ with $\pi(x)$ in \eqref{eq:system} and
integrating the resulting ODE on the interval $\intcc{0}{\tf}$.



\subsection{HJB Numerical Discretization}
\label{sec:FMM}

\topic{no closed-form solution}

When an analytical solution is unavailable for \eqref{eq:hjb}, we must
resort to numerical PDE solvers. To this end, we follow closely the
FMM discretization~\tocite{Sethian99}.

\topic{Definitions}

\begin{definition}[Sample Set]
  A finite or countable set of distinct vertices $\Xd$
  is called a \emph{sample set} of $X$.
\end{definition}

\begin{definition}[Abstract and Geometric Simplices]
  An \emph{abstract simplex} of dimension $d$ is a set of $(d+1)$
  vertices $\simplex = \set{x_0, \ldots, x_{d}} \subset \Xd$ such that
  $x_k \not= x_{k'}$ for all $k$ and $k'$.
  A \emph{geometric realization} of $\simplex$ is a convex hull of its
  vertices, which we call \emph{geometric simplex} and denote
  $X_\simplex$. Formally,
  \begin{equation}
    X_\simplex =
    \Bigcset{\sum^{d}_{k = 0} \alpha_k x_{k}}{\forall \alpha_k \ge 0 \;\; \text{such that} \;\; \sum^{d}_{k = 0} \alpha_k
      = 1} \prd 
  \end{equation}
  In the above, $\set{\alpha_{k}}^d_{k=0}$ are called
  \emph{barycentric coordinates} of a point $x = \sum^{d}_{k =
    0} \alpha_k x_{k} \in X_\simplex$. Finally, $X_{\simplex'}$ is
  called a \emph{(proper) facet} of $X_\simplex$ if $\simplex'$ is a
  (proper) subset of $\simplex$.
\end{definition}

\begin{definition}[Vertex Set Triangulation]
  A triangulation of $\Xd$ is a set of abstract $d$-dimensional
  simplices $\simcom$ such that for all $\simplex,
  \simplex' \in \simcom$ and $\simplex \not= \simplex'$ the
  intersection $X_{\simplex} \bigcap X_{\simplex'}$ is a proper facet
  of both simplices and the union $\bigcup_{\simplex \in
    \simcom} X_\simplex$ covers the convex hull of $\Xd$.
\end{definition}

\topic{Approximate discretization of $\Xfree$}



The numerical solution is computed in three steps.  

First, we consider a triangulation of a sample set of vertices in
$X$. An approximation of $\Xfree$ is then derived from this vertex set
triangulation by ignoring simplices that have a nonempty intersection
with $\Xobs$. Note that this approximation approaches $\Xfree$ in the
limit, as the dispersion of the vertex set tends to zero.



\topic{Cost-to-go approximation}


Second, using a collision-free 
triangulation of $\Xfree$, we define a piecewise linear cost-to-go
approximation:
\begin{equation}
  \label{eq:pwlinear}
  \hat V(x) = \sum_{x_k \in \simplex} \hat V(x_k) \alpha_k(x) \cma
\end{equation}
in which $\simplex$ is an abstract simplex such that $x \in
X_\simplex$%
, and $\alpha_k(x)$ is $k$th barycentric coordinate of $x$ in
$X_\simplex$.  Note that $\hat V$ is defined completely using its
values at sampled vertices.

\topic{Discrete dynamic programming}

Finally, we substitute \eqref{eq:pwlinear} into \eqref{eq:hjb} and
derive the discrete \HJB equation at vertex $x \in \Xd$:
\begin{equation}
  \label{eq:discrete_hjb}
  \min_{\simplex \in \St(x)}\min_{u\in U_{x,\simplex}} \set{\pair{\nabla_\simplex \hat V(x)}{f(x,u)} + c(x,u)} = 0 \prd
\end{equation}
In the above, $\nabla_\simplex \hat V$ is a restriction of $\nabla
\hat V$ onto simplex $X_\simplex$ and $\St(x) = \cset{\simplex}{x \in
  \simplex}$ is called a \emph{star} of vertex $x$ and represents a
local neighborhood around $x$. The minimization over input signal $u$
is constrained to the set $U_{x,\simplex} = \cset{u \in U}{\exists
  \delta > 0 : x + \delta f(x, u) \in X_\simplex}$. This constraint is
necessary to construct a \emph{positive coefficient}
discretization, which converges to the viscosity solution of
\eqref{eq:hjb}~\tocite{CraLio83}.

\topic{Convergence rate}

The linear convergence rate of the numerical
discretization~\eqref{eq:discrete_hjb} has been established in
\tocite{YerLav12}, that is, $E \le C h$, in which $E = \sup_{x \in X}
\abs{\hat V(x) - V(x)}$ is the global numerical error, and $h =
\sup_{\simplex \in \simcom} \max_{x,x' \in \simplex} \norm{x - x'}$ is
the maximum edge length. In the above, $C > 0$ is a constant, which is
independent of $h$.

\topic{Monotonically increasing cost-to-go and the Fast Marching Method}

When considered at all vertices $x \in \Xd$,
equation~\eqref{eq:discrete_hjb} defines a system of nonlinear
equations. Similar to Dijkstra's algorithm, the Fast Marching Methods
(FMM)~\tocite{Sethian99} evaluates the cost-to-go values in a single
pass trough the vertex set using a priority queue. If the
triangulation is acute, then the FMM solves this nonlinear
system~\tocite{KimSet98,SetVla00}.


\subsection{Delaunay Triangulation}
\label{sec:DT}

\begin{definition}[Delaunay Triangulation]
  The Delaunay (also known as Delone) triangulation, $\simcom$, of a
  sample set $\Xd$ is such that, for all $\simplex \in \simcom$ and
  all $x \in \Xd \setminus \simplex$, $x$ is located outside of the
  circumsphere of $X_\simplex$.
\end{definition}



\topic{Relation to convex hulls. The existence and uniqueness of the
  DT}

General vertex set triangulations are difficult to compute, with the
notable exception of the Delaunay triangulation (DT). In computational
geometry, the bijection between Delaunay simplices in $\Re^d$ and
lower faces of the convex hull of sample vertices lifted onto a
paraboloid in $\Re^{d+1}$ has been established. Thus, geometric convex
hull algorithms can be used for constructing the DT. It also follows
from this relation that the DT exists and is unique if sample vertices
are in \emph{general position} (that is, there are no $d+2$ vertices
such that they belong to some $d$-dimensional sphere).

Moreover, compared with other vertex set triangulations, the DT
maximizes the min-containment radius of simplices~\tocite{Rajan94} and
minimizes the second-order error of a piecewise linear interpolation
in 2D~\tocite{BerEpp92} and in $d$-dimensional
case~\tocite{CheXu04}. The former implies that the DT is the most
regular of all triangulations, and the latter guarantees the smallest
possible numerical error in \eqref{eq:discrete_hjb}.

Considering the
simplicity of computing the DT and its optimality with respect to
interpolation error, we use the DT for approximating the cost-to-go
function on $\Xfree$.

\subsection{Stochastic Delaunay Triangulation}
\label{sec:SDT}


\topic{random sample motivation}

Random sampling has been proven useful for high-dimensional path
planning problems. Randomized motion planning algorithms have both
theoretical and practical advantages. In theory, they are
probabilistically complete, which guarantees finding an existent
solution. In practice, they rapidly explore the environment searching
for a solution.

\topic{Stochastic DT}

Following this trend, we present the Delaunay triangulation of a
random vertex set, which we call \emph{stochastic Delaunay
  triangulation} (SDT). From the motion planning perspective, the SDT
can be compared with the PRM in that they both approximate the free
space, as the number of sample vertices increases. Unlike the PRM
graph, however, the SDT is a volumetric approximation of
$\Xfree$. Although computing simplex collisions is generally more
difficult than performing an $1$-dimensional collision check, this
volumetric information enables finding paths that traverse through
simplicial cells instead of being constrained on graph edges.



%

\topic{Poisson-Delaunay mosaics}

In this section, we establish a connection between the SDT and
Poisson-Delaunay mosaics and some useful properties of the former that
follow from the integral geometry.

\topic{Poisson point process}

\begin{definition}[Poisson Point Process in $\Re^d$]\label{def:PPP}
  A Poisson point process in $\Re^d$ of intensity $\rho$ is a set of
  random points such that for two bounded, measurable, disjoint sets
  $X_1$ and $X_2$, the number of points of this process that are
  inside $X_1$ and $X_2$ are two independent Poisson random variables
  with parameter $\rho \mu(X_1)$ and $\rho \mu(X_2)$,
  respectively.\footnote{Here, $\mu(X)$ denotes the \Lebeg measure of
    $X$.}
\end{definition}

\begin{definition}[Poisson-Delaunay Mosaic]\label{def:PDM}
  Let $\Xd$ be a realization of a Poisson process. The Delaunay
  triangulation of $\Xd$ is a Poisson-Delaunay mosaic.\footnote{The
    definition of DT can be extended to locally sparse infinite sets.}
\end{definition}

\topic{Poisson point process and uniform sampling}

It follows from \defref{def:PPP} that a set of $N$ uniformly and
independently distributed vertices from the free space is the
restriction of the Poisson point process. Using maximum likelihood
estimate, we establish the intensity of this process: $\rho = N /
\mu(\Xfree)$. This relation shows that the SDT can be considered as
the restriction of a Poisson-Delaunay mosaic.


\topic{Properties of Poisson-Delaunay mosaics}


In integral geometry, statistical properties of Poisson-Delaunay
mosaics have been established. For example, the expected number of
simplices in the star of a vertex depends on the dimension number $d$,
but it is independent of the intensity
$\rho$~\tocite{Kendall90}. Therefore, the expected number of
vertex-adjacent simplices is independent of $N$ for the
SDT.\footnote{Note that our optimal planner has a constant branching
  factor in expectation compared with $O(\log N)$ branching factor for
  \RRTstar.}


\topic{Statistics on the shape and size of a DT simplex}

Another property of Poisson-Delaunay mosaics is related to the average
simplex size as a function of the point process intensity. The average
edge length is $\Theta(\rho^{-1/d})$~\tocite{Miles74}, and the
expected maximum edge length is $\Theta((\log \rho/
\rho)^{1/d})$~\tocite{BerEppyao91}. Using previously derived relation
between $\rho$ and $N$, we find that the expected edge length and its
maximum of the SDT are $\Theta(N^{-1/d})$ and $\Theta((\log N/
N)^{1/d})$, respectively.\footnote{Note that exactly the same bound on
  the shrinking radius enables percolation limit in the RRG-based
  algorithms.}

\section{SAMPLING-BASED FEEDBACK PLANNING ALGORITHM}
\label{sec:algorithm}

\begin{figure*}[t]
  \footnotesize

  \includegraphics[height=15.25cm, trim=127 205 245 50, clip=true]{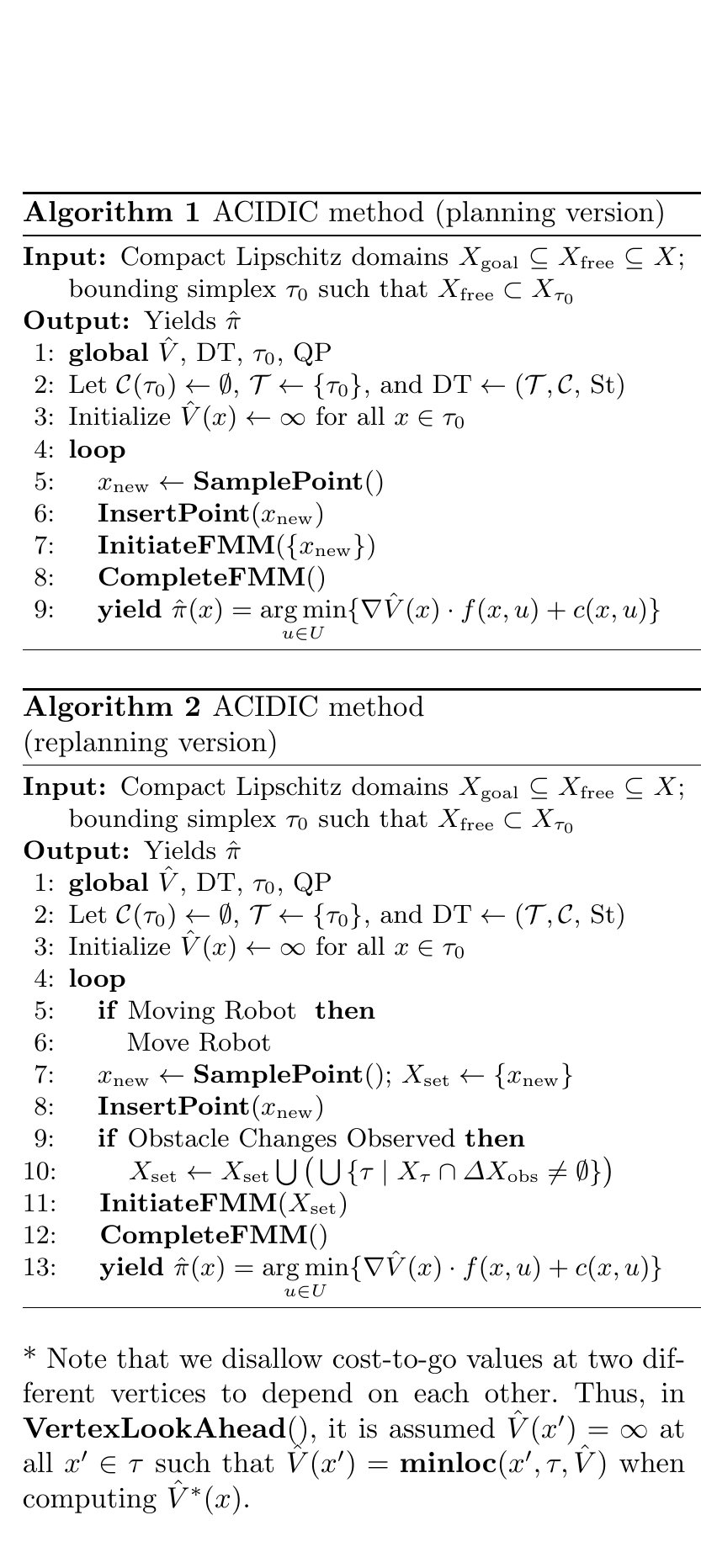}
  \includegraphics[height=15.25cm, trim=127 205 295 50, clip=true]{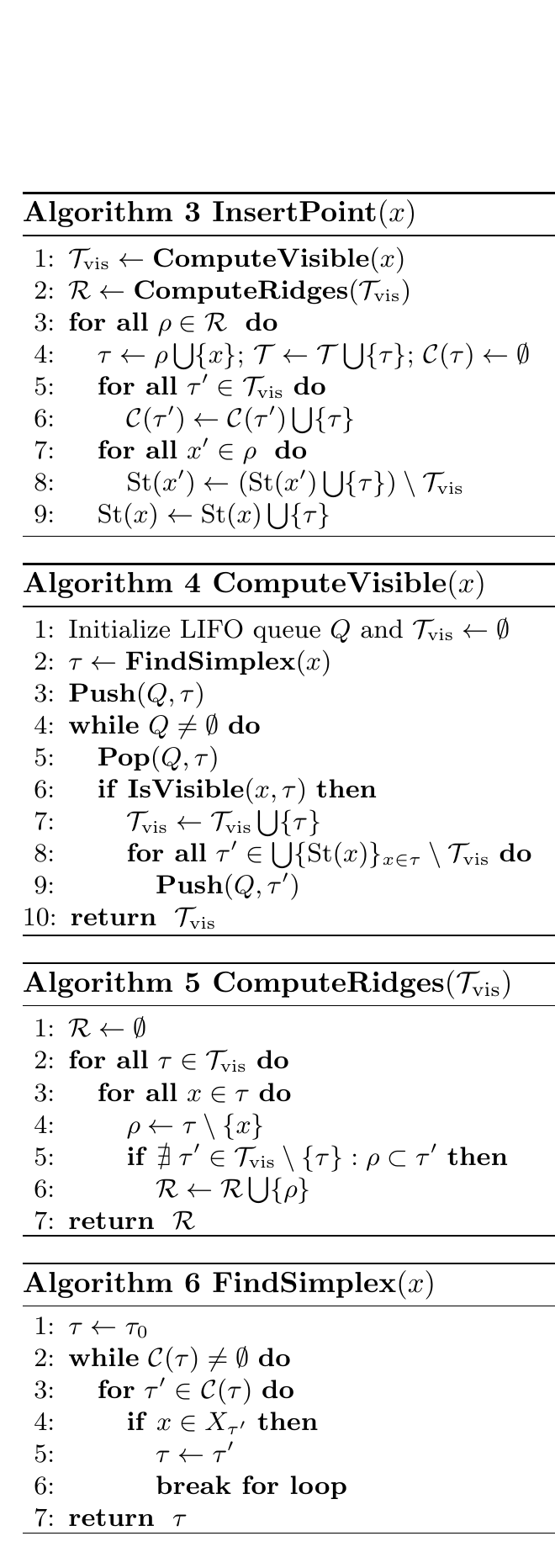}
  \includegraphics[height=15.25cm, trim=127 205 295 50, clip=true]{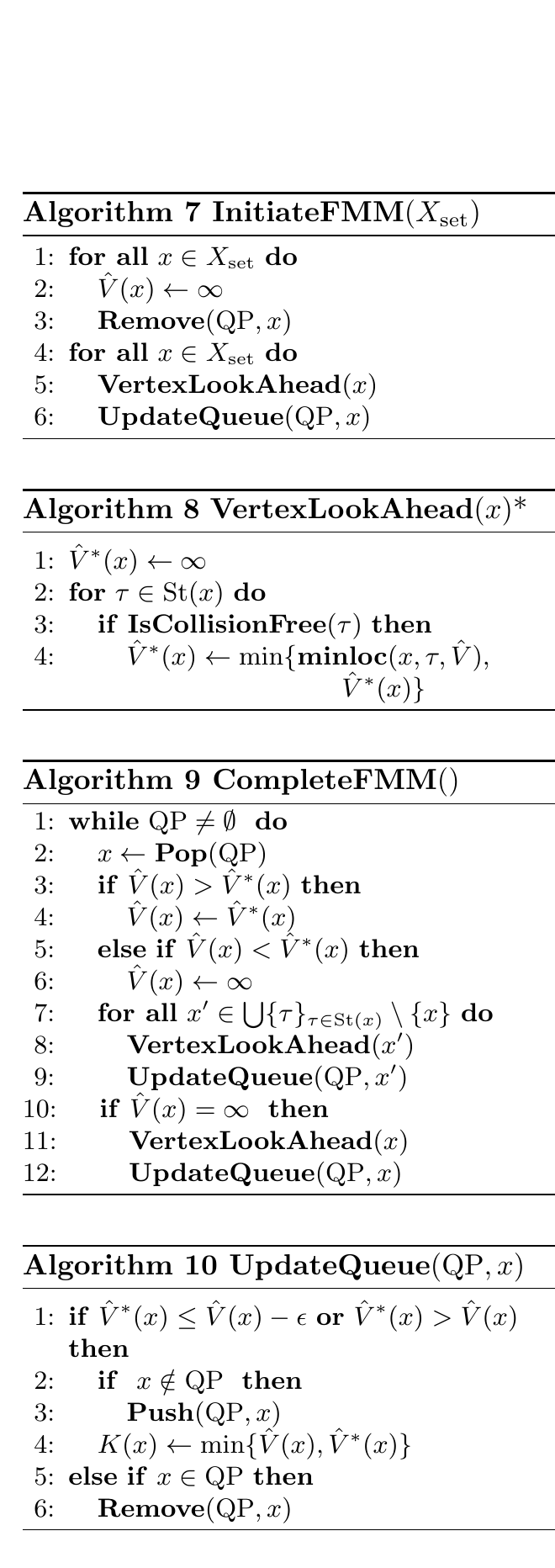}

\end{figure*}

%

\topic{Steps of sampling-based feedback planning}

We now present the ACIDIC method for sampling-based feedback
planning. The execution trace of our method is similar to most
sampling-based path-centric planners:
\begin{enumerate}
\item Sample a new vertex $\xRand$ from $X$%
;
\item Refine the Delaunay triangulation to include $\xRand$%
;
\item Update the cost-to-go values and associated feedback control in
  the simplicial approximation of $\Xfree$%
.
\end{enumerate}

\topic{high-level description of ACIDIC}

At a conceptual level, the ACIDIC method ``etches'' the free space
away from the obstacle set while simultaneously refining a feedback
control. 
In the limit of infinitely many sampled vertices, all points of
$\Xfree$ become part of the triangulation and the optimal feedback
control is computed.

\topic{family of algorithms}

This basic idea can be used to create an entire family of feedback
planning algorithms. The behavior of a particular algorithm depends on
three factors: 
the sampling strategy, 
the volumetric decomposition (that also includes collision-detection
strategy), and 
the methods by which the approximate cost-to-go function and 
feedback control are updated%
.

\topic{family of algorithms}

We discuss our implementation of these three steps in
Sections~\ref{sec:sampling}--\ref{sec:update_cost}, respectively. An
illustrative example is presented in \secref{sec:exampleAlg}, and an
extension to replanning in dynamic environments in
\secref{sec:replanning}. Our generic implementation can be improved
for specific applications.

\subsection{Sampling Strategy} \label{sec:sampling}

\topic{sampling strategies}

Motivated by theoretical results from \secref{sec:prelim}%
, we explored three different sampling strategies, which 
 we implemented in 
\algref{alg:AOFP}:
uniform random sampling, 
deterministic largest-simplex Delaunay refinement, and 
goal-oriented refinement in the vicinity of the current shortest path
and near obstacle boundaries.

\topic{uniform sampling}

In theory, uniform random sampling guaranties regularity and
convergence of the DT in expectation. In practice, however, biased or
carefully engineered deterministic sequences are likely to improve
planning algorithm performance by constructing high-quality
triangulations.




\topic{Delaunay refinement}

The largest circumradius of all Delaunay simplices defines the
Euclidean dispersion of the sampled vertex set. Moreover, the
circumcenter of the corresponding simplex maximizes the distance to
the closest sample vertex. By placing $\xRand$ at this circumcenter,
we optimize the sample sequence to consider unexplored regions of the
configuration space.
This strategy is known as Delaunay refinement~\tocite{Chew89}, and is
proven to produce a high-quality triangulation~\tocite{Shewchuk02}.

\topic{Goal-oriented refinement} In the limit, Delaunay refinement
samples vertices uniformly in the configurations space. However, the
convergence of our algorithm can be improved by sampling regions in
which the interpolation error is the highest. We consider, for
example, regions where the cost function is highly nonlinear, such as
near obstacle boundaries or the goal set as well as the vicinity of
the current optimal path.
%
In particular, we choose $\xRand$ as the circumcenter of the largest
circumsphere of Delaunay simplices that are either on the boundary of
$\Xobs$ and $\Xgoal$ or contain the current shortest path.



\subsection{Computing Volumetric Free Space Approximation} \label{sec:collision_detection}

At every iteration, \algref{alg:DTUpdate} uses a newly sampled vertex
to improve the volumetric approximation of $\Xfree$. To update the
Delaunay triangulation, we follow closely the incremental convex hull
algorithm introduced in~\tocite{BaDoHu96}. To this end, we find all
simplices that are \emph{visible} from the newly inserted vertex; see
\algref{alg:ComputeVisible}. Next, we find a set of \emph{ridges},
which are defined as faces separating visible and invisible simplices;
see \algref{alg:ComputeRidgesValleys}. A set of new Delaunay simplices
is created by connecting all ridges with the inserted vertex; see
\linref{alg:DTUpdate:CreateSimplex} in \algref{alg:DTUpdate}. Finally,
we ``remove'' visible simplices by updating their children sets to
include all newly inserted simplices. \todo{An illustration of this
  procedure is depicted in \figref{fig:}}


Note that all simplices are actually retained; however, only childless
simplices belong to the current DT, which is reflected in updating the
local connectivity information in $\St(x)$ and $\St(x')$; see
Lines~\noHyperLink{\ref{alg:DTUpdate:NeighborLoop}}--\noHyperLink{\ref{alg:DTUpdate:UpdateStar}}
in \algref{alg:DTUpdate}. The remaining simplices organized into a
directed acyclic graph structure that helps locating future vertices
within the Delaunay triangulation, as it is prescribed by
\algref{alg:FindSimplex}.

After the DT is updated, a black box collision-detection module is
used to find free-to-traverse simplices in the current
triangulation. We assume the conservative collision-detection
implementation, that is, the simplex is considered collision-free if
$X_\simplex \cap \Xobs = \emptyset$.


It is worth noting that pointwise collision-detection modules are
ill-suited for volumetric representations because each simplex
contains an infinite number of points to check. Thus, as it is also
the case with many 
graph-based planning algorithms, we require additional information
about the obstacle set to compute the volumetric free space
approximation. For example, if the distance to the nearest obstacle is
known and higher-order dynamics, such as velocities, are bounded, then
we can verify collision-free simplices by solving a convex
minimization problem.
%
%
Volumetric collision-detection is an advanced topic, and further
implementation details are beyond the scope of the current paper.

\subsection{Computing Cost-To-Go Function and Feedback Control} \label{sec:update_cost}


The cost-to-go function and the optimal feedback control are
maintained using a modified version of the Fast Marching Method (FMM).
In particular, our version addresses a nonmonotonic cost-to-go
wavefront propagation, which may be caused by either a nonacute
triangulation or a part of the free space approximation becoming the
obstacle set due to local simplex rewiring and a conservative
collision-detection.


Our modifications to the FMM follow closely replanning path-centric
strategies, such as \Dstar Lite~\tocite{KoeLik02} and
\RRTx\tocite{OttFra14}, which deal with appearing and disappearing
obstacles by propagating the \emph{increase} wavefront ahead of the
\emph{decrease} wavefront. To prevent infinite loops, the algorithm
interrupts wavefront propagation when changes are smaller than a given
parameter~$\epsilon$. At the end of this section, we investigate a
peculiar side effect of our implementation: a straightforward
extension to an optimal feedback replanning algorithm.


\subsubsection{Fast Marching Feedback Planning Algorithm} \label{sec:exampleAlg}

The ACIDIC planning algorithm (\algref{alg:AOFP}) improves
incrementally the feedback control by initiating and propagating the
cost-to-go wavefront (Lines~\noHyperLink{\ref{alg:AOFP:extend}}
and~\noHyperLink{\ref{alg:AOFP:reduce}}, respectively).

The wavefront is initiated at newly sampled vertex $\xRand$, when the
look-ahead value $\lookAheadCost$ becomes inconsistent with the
current cost-to-go value $\cost$; see \algref{alg:extendRRTSharp}. The
look-ahead value is computed in \algref{alg:wire} using $\minloc$
function that solves the inner minimization problem
in~\eqref{eq:discrete_hjb} for collision-free simplices. The
implementation of $\minloc$ and its geometric interpretation are
discussed 
in~\tocite{YerLav12}. Finally, we also assume $\minloc$ returns
$0$ if $x$ is in the goal set.

The inconsistency between $\cost$ and $\lookAheadCost$ implies that
the solution of the nonlinear system is not found
yet. \algref{alg:reduceRRTSharp} repairs inconsistency
by propagating the cost-to-go wavefront through the environment in the
fast marching fashion. To this end, the vertices are organized in a
priority queue in the increasing order of their key values $\key =
\min\set{\cost, \lookAheadCost}$. For each vertex $x$, two cases are
considered: 1)~cost-to-go value decreases 
(\linref{alg:reduceRRTSharp:decrease})
 and 2)~cost-to-go value increases 
(\linref{alg:reduceRRTSharp:increase}).
In the first case, $\costOf{x}$ is updated to its current best
estimate, $\lookAheadCostOf{x}$, and the decrease wavefront is
propagated to vertex neighbors lowering their $\cost$ values.
In the second case, $\costOf{x}$ is set temporarily to infinity
causing $\lookAheadCost$ values to increase at all neighbors whose
cost-to-go values are depended on $\costOf{x}$. Thus, the increase
wavefront is propagated. Next, $\lookAheadCostOf{x}$ is updated
creating the decrease wavefront, which repairs inconsistencies
introduced by the increase wavefront.

When \algref{alg:reduceRRTSharp} terminates, the cost-to-go function
is consistent with \eqref{eq:discrete_hjb} up to a small $\epsilon$
error in the entire domain. However, if the initial position is known
the global cost-to-go computations are unnecessary, and the estimate
of the cost-to-come can be used to restrict the computation
domain. Various heuristics and their effect on computations are
discussed in~\tocite{ClaChaVla13,YerLav12}.

\subsubsection{Fast Marching Replanning Algorithm} \label{sec:replanning}

The goal of the ACIDIC replanning algorithm is to update the feedback
control as soon as robot's sensors detect a change in obstacle
configuration. Fast control loop relies crucially on replanning
algorithm efficiency. In this case obstacles may appear or disappear,
which results in increasing or decreasing cost-to-go values.
Fortunately, ACIDIC planning algorithm accounts for both of these
changes, and its extension towards replanning algorithm is rather
straightforward.


We present \algref{alg:AOFRP}, in which wavefront propagation is
initiated once the change in obstacle set is confirmed
(\linref{alg:AOFRP:inconsistent}). A volumetric collision-detection
module is used to find all inconsistent vertices that are affected by
obstacle changes. This conservative estimate guarantees that after
wavefront propagation the computed cost-to-go function is consistent
with \eqref{eq:discrete_hjb} up to a small $\epsilon$ error.

\section{ANALYSIS of the ALGORITHM}

\subsection{Numerical Convergence}

The resolution of the sample set increases when each additional vertex
is inserted. Hence, the accuracy of the planning algorithm is expected
to improve. This intuition can be rigorously supported combining the
results of \secref{sec:prelim}.

\begin{theorem}[Numerical Convergence]
  We assume $\epsilon = 0$ and $N$ random vertices are sampled. The
  expected solution error, $\Exp[\Error]$, is then $O((\frac{\log
    N}{N})^{\frac1d})$.
\end{theorem}
\begin{proof}
  The proof follows from the numerical error bound presented in
  \secref{sec:FMM} and the expected maximum edge length presented in
  \secref{sec:SDT}.
\end{proof}



\subsection{Computational Complexity}

To establish the computational complexity of \algref{alg:AOFP}, we
consider its expected runtime per iteration.

\topic{Output complexity and efficient incremental algorithms}

It should be noted that the worst-case runtime of our algorithm is
bounded from below by the maximum number of Delaunay simplices in the
triangulation of $N$ vertices, which is proportional to
$N^{\ceil{d/2}}$~\tocite{McMullen70}. However, such artificially
constructed cases ``rarely'' occur in practice. Thus, algorithms that
we proposed for vertex insertion and wavefront propagation are
\emph{output-sensitive} in that they have optimal complexity $O(\log
N)$ per Delaunay simplex. In conjunction with the results from
\secref{sec:SDT} we prove the expected runtime bounds for the ACIDIC
method.

\begin{lemma}\label{lem:near_visible}
  Let $\simcom$ be a DT of a random vertex set $\Xd$, and let $\xRand
  \in X$.  For the closest vertex $x^* = \argmin_{x \in \Xd}
  \norm{\xRand - x}$, there exists $\simplex \in \St(x^*)$ such that
  $x$ is inside the circumsphere of $X_\simplex$.
\end{lemma}
\begin{proof}
  We omit the proof due to space limitations.
\end{proof}
\begin{theorem}[Delaunay Triangulation Complexity]
  The expected runtime of \algref{alg:DTUpdate} is $O(\log N)$ per
  iteration.
\end{theorem}
\begin{proof}
  Since the expected size of $\St(x)$ is independent of $N$, the
  visible simplex number and the ridge number are constant in
  expectation. Thus, average simplex rewiring runtime is
  constant. Moreover, from \lemref{lem:near_visible}, it follows that
  finding the first visible simplex can be done in expected $O(\log
  N)$ time using Kd-trees~\tocite{FriBenFin77}.
\end{proof}

Note that our algorithm for finding the first visible simplex avoids
using Kd-trees for simplicity. In theory, it is not yet clear that the
average depth of the children graph is $O(\log N)$. However, this
algorithm performs well in practice.




\begin{theorem}[Wavefront Propagation Complexity]
  The expected runtime of \algref{alg:reduceRRTSharp} is $O(\log
  N)$ for all $\epsilon > 0$.
\end{theorem}
\begin{proof}
  Since the approximate cost-to-go function converges, the number of
  $\epsilon$-changes of $\cost$ at each vertex is bounded. Thus, the
  number of times wavefront propagates through each vertex is
  constant. Maintaining the priority queue, however, takes $O(\log N)$
  time.
\end{proof}
   

\section{EXPERIMENTS, RESULTS, and DISCUSSION}


The basic idea presented in this paper can be used to create an entire
family of incremental volumetric feedback planning algorithms. For
brevity, we limit our investigation to the implementations proposed in
\secref{sec:algorithm}.
In Sections~\ref{sec:planning2D} and \ref{sec:planningND} we
experimentally evaluate the performance of feedback planning in static
environments and compare our results with state-of-the-art graph-based
methods.
In \secref{sec:planningExperiments} we present simulations of the
replanning version of the algorithm in a dynamic environment.


All experiments were run in simulated environments on a Dell Optiplex
790 with Intel i7 chip and 16GB of RAM; a single processor core is
used for all experiments. All algorithms are implemented in Julia
programming language. Both volumetric and graph-based algorithms
use the same code-base in order to make the comparison as fair as
possible. For example, all algorithms use the same sampling schemes,
obstacle representations, collision-detection routines, heap
data-structures, control-loop computations, data-logging procedures,
and so on.

\subsection{Point Robot in Random 2D Static
  Environment} \label{sec:planning2D}

\begin{figure}[t]
  \footnotesize

  \vspace{0.1cm}

  \hspace{0.2cm} Random Sampling \hspace{0.4cm} Delaunay Refinement \hspace{0.4cm} Focused Refinement

  \includegraphics[width=2.8cm, trim=160 270 250 300, clip=true]{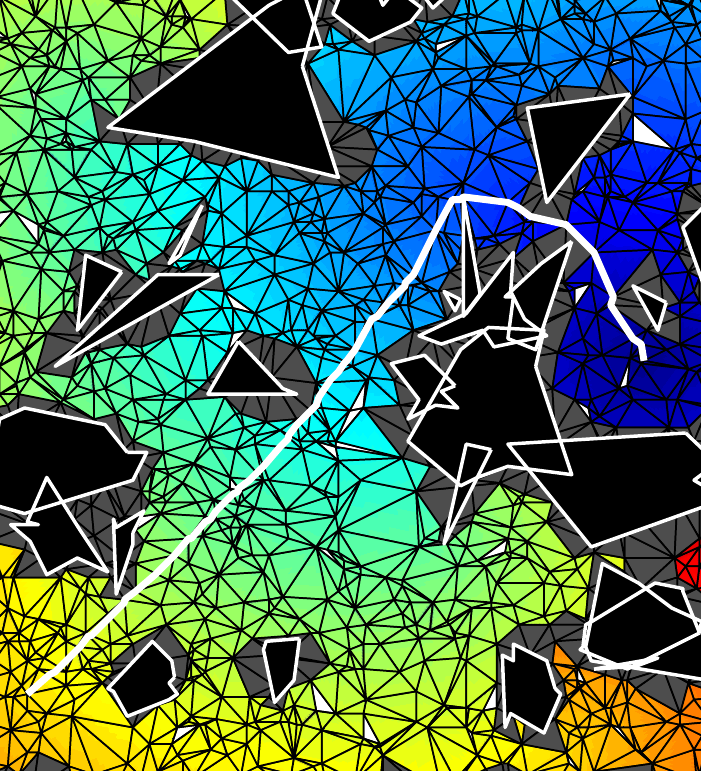}
  \includegraphics[width=2.8cm, trim=160 270 250 300, clip=true]{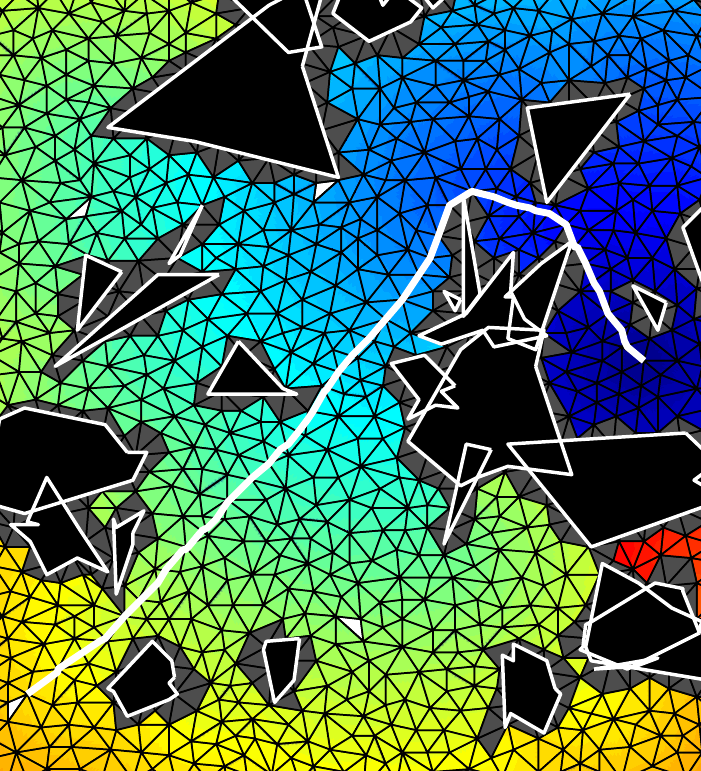}
  \includegraphics[width=2.8cm, trim=160 270 250 300, clip=true]{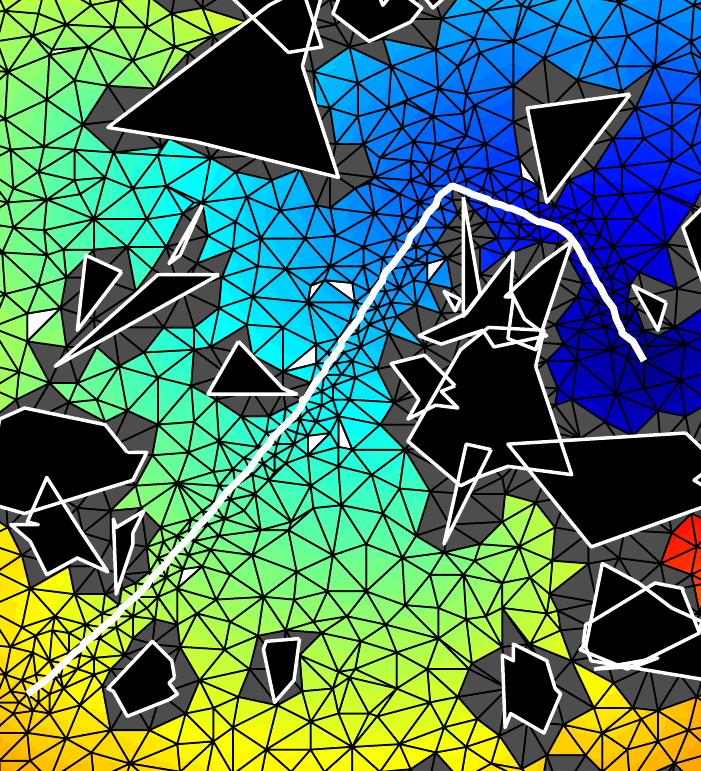}

  \caption[]{Temperature map of the cost-to-go values (color) and path
    (white) computed by ACIDIC algorithm during the first 10 seconds
    in a randomly generated 2D environment of size $100\times100$.
    Delaunay simplices are outlined in black, and obstacles are solid
    black.}


  \label{fig:environment1}
\end{figure}

\begin{figure}[t]
  \footnotesize

  \centering

  \includegraphics[width=4.2cm, trim=200 295 215 290, clip=true]{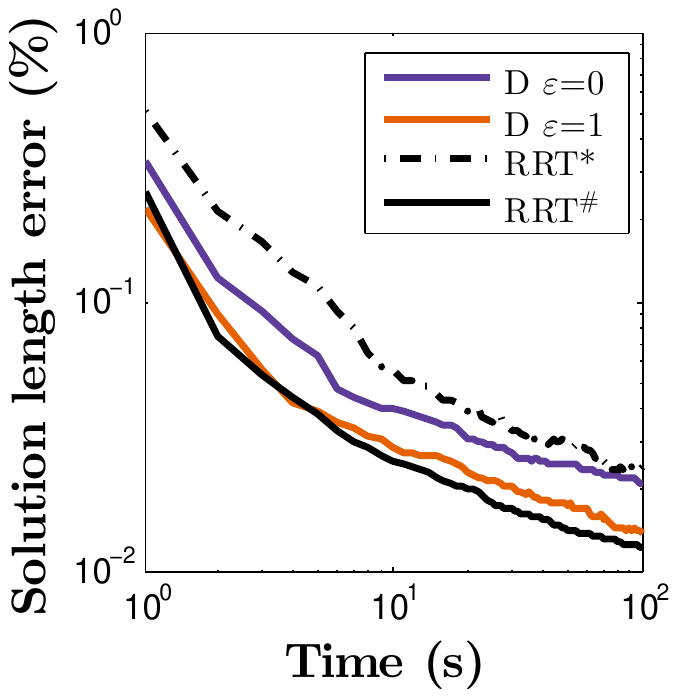}
  \includegraphics[width=4.2cm, trim=200 295 215 290, clip=true]{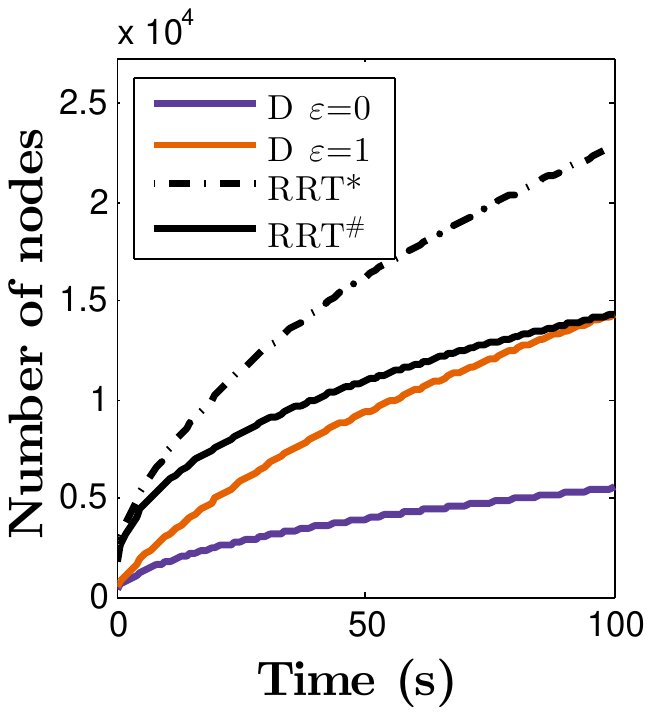}

  \caption[]{Averaged over 50 trials relative path error (left,
    log-log scale) and the vertex number (right, regular scale) against
    wall time for ACIDIC algorithm using Delaunay refinement with
    $\epsilon = 0$ (purple) and $\epsilon = 1$ (orange), \RRTstar
    (black dot-dash), and \RRTsharp (black solid) in the environment
    of \figref{fig:environment1}.}
  


  \label{fig:experiment1}
\end{figure}

\begin{figure}[t]
  \footnotesize

  \centering

  \includegraphics[width=4.2cm, trim=200 295 200 280, clip=true]{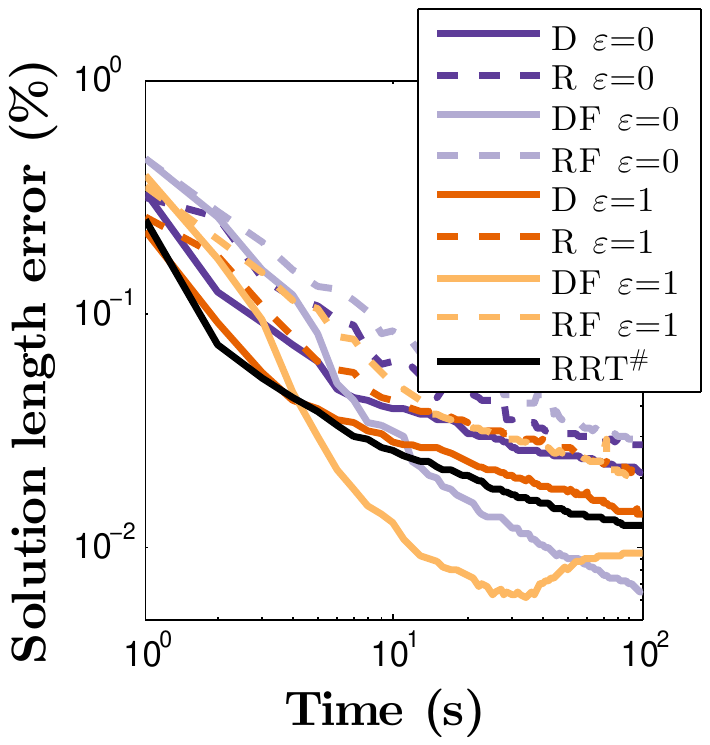}
  \includegraphics[width=4.2cm, trim=200 295 200 280, clip=true]{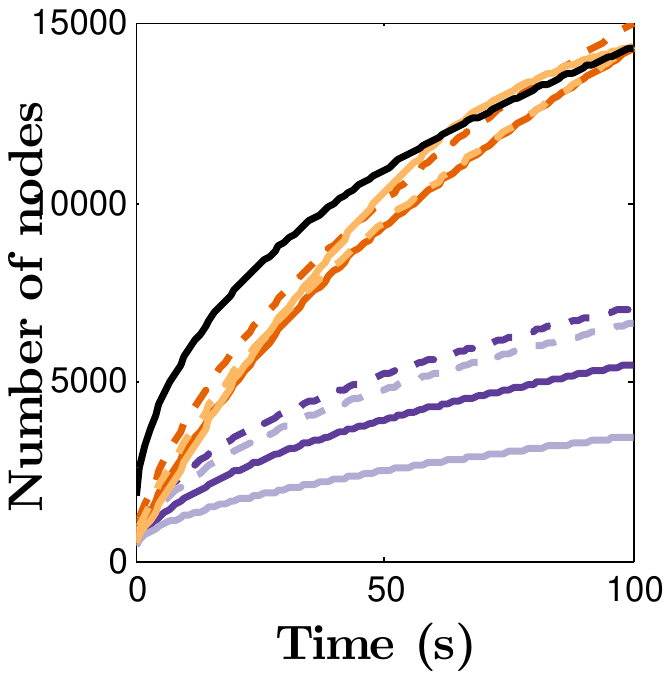}

  \caption[]{Averaged over 50 trials relative path error (left,
    log-log scale) and the vertex number (right, regular scale) against
    wall time in the environment of \figref{fig:environment1}.
    Results are for ACIDIC algorithm using Delaunay refinement (solid)
    and random sampling (dashed), $\epsilon = 0$ (dark/light purple)
    and $\epsilon = 1$ (orange/yellow), uniform (dark purple/orange)
    and focused (light purple/yellow) sampling, and \RRTsharp
    (black).}


  \label{fig:experiment1B}
\end{figure}

In this experiment, a holonomic single integrator point-robot desires
to travel to a goal point among randomly generated polygonal obstacles
in 2D; see \figref{fig:environment1}.
%
%
We compare 
our ACIDIC planning algorithm
with \RRTstar and \RRTsharp. Results of most basic versions appear in
\figref{fig:experiment1}, while a comparison of variants that use
different sampling techniques and values of $\epsilon$ appears in
\figref{fig:experiment1B}.
%

In theory, volumetric and graph-based methods have identical
asymptotic performance per iteration. In experiments, however, we
observe that volumetric methods have a constant factor overhead
associated with updating the Delaunay triangulation and computing fast
marching wavefront propagation. Thus, our method processes fewer
vertices per unit of wall time. Despite the difference in sampling
performance, convergence rates of all considered methods are
comparable, which implies that the ACIDIC method is more efficient in
using sampled vertices. We attribute this efficiency to allowing the
path traverse through the volume of space potentially ignoring all
sampled vertices.

We optimized the convergence rate of the ACIDIC method using only the
information gained from the volumetric decomposition. First, refining
near obstacle boundaries is accomplished by sampling simplices that
are partially in collision. This focused sampling enables the
discovery of narrow corridors in the free space. Hence, the optimal
solution can be found using fewer vertices. Second, the convergence
rate is improved by refining the triangulation in the vicinity of the
best path. Finally, truncating wavefront propagation reduces the
runtime per iteration. The optimized ACIDIC method outperforms all
graph-based algorithms considered in our experiments.

Note that truncating the wavefront propagation by setting $\epsilon =
1$ significantly improved the convergence in static environments.  In
graph-based methods, the usefulness of such this idea appears to be
limited to dynamic environments, in which fast wavefront propagation
and the control-loop speed are primary concerns. We believe the
reason why the truncation benefits the convergence of ACIDIC methods
is due to the larger constant factor associated with recomputing the
cost-to-go function through the triangulation.


%

\subsection{Block Obstacle in $\Re^d$} \label{sec:planningND}

\begin{figure}[t]
  \footnotesize

  \centering

  \includegraphics[width=3.2cm, trim=227 295 235 295, clip=true]{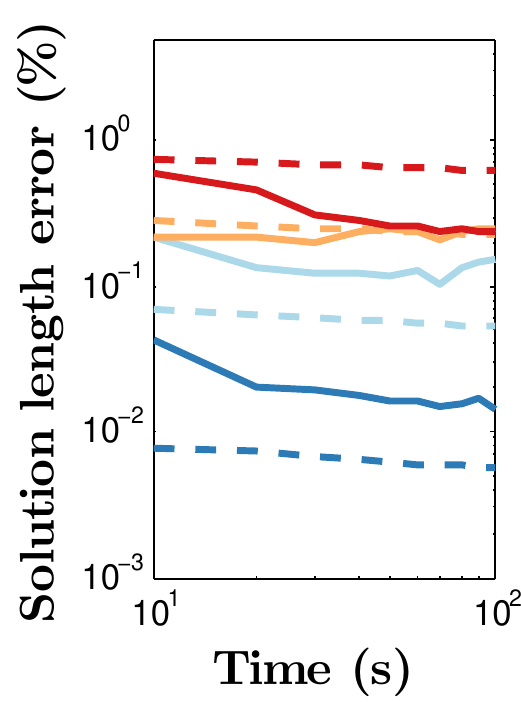}
  \includegraphics[width=2cm, trim=260 260 260 320, clip=true]{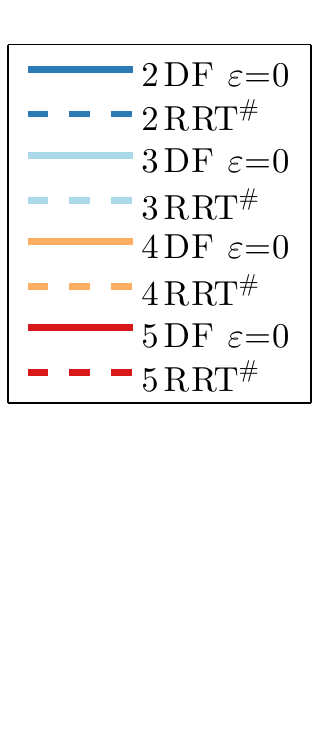}
  \includegraphics[width=3.2cm, trim=227 295 235 295, clip=true]{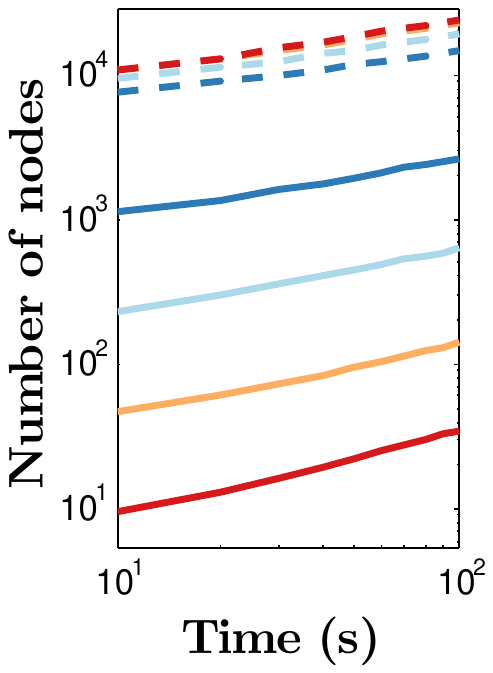}

  \caption[]{Averaged over 50 trials relative path error (left,
    log-log scale) and the vertex number (right, log-log scale) against
    wall time for ACIDIC algorithm (solid) 
    and \RRTsharp (dashed) in 2D to 5D environments (color).}


  \label{fig:experiment2}
\end{figure}

In \figref{fig:experiment2}, we evaluate the performance of ACIDIC
method in dimensions between $2$ and $5$. The environment used for
each dimension is a hypercube with a single prismatic obstacle located
at the center. The obstacle spans one-half of the environment in the
first two dimensions and has infinite length along other dimensions.

As with many planning methods, the volumetric idea suffers from the
``curse of dimensionality''. From the experiments, it became evident
that the runtime per vertex grows exponentially with the dimension
number for the ACIDIC algorithm. Note that this overhead is constant
in any particular dimension, and the expected vertex insertion
complexity for ACIDIC algorithm remains $O(\log N)$, in which $N$ is
the number of already inserted vertices. Opposite to the volumetric
methods, the runtime per vertex is largely independent of the
dimension number for graph-based algorithms, which, unfortunately,
does not yield their faster convergence. From the convergence results,
we confirm that volumetric methods use vertices more efficiently. We
attribute this to the fact that $d+1$ vertices are sufficient to
construct a simplex covering large volume of the free space. On the
other hand, graph-based methods may require many more sample vertices
to recover the shortest path.


\subsection{Replanning Simulations in Dynamic Environments} \label{sec:planningExperiments}

Due to space limitations and also the fact that navigation through
dynamic environment is best visualized in media that incorporates
time, we have uploaded a playlist of movies to
\url{http://tinyurl.com/qjnazvr} that illustrate the dynamic version
of ACIDIC replanning algorithm.

The replanning version of ACIDIC is the first asymptotically-optimal
sampling-based feedback replanner that is capable of repairing the
feedback control in real-time when obstacles unexpectedly appear,
disappear, or move within the configuration space.
While \RRTx is closely related to our algorithm, it is path-centric and
requires a feedback control to be computed in post-processing.  While
this can be done quickly, in practice, the particular method that is
used will significantly affect reaction time --- the most important
performance measure of a replanning algorithm. Thus, we do not compare
the two methods directly.

\section{SUMMARY}

Proposed in this paper, is the first asymptotically optimal feedback
planning algorithm that uses a volumetric approximation of free space
in conjunction with the Fast Marching Method.
The main idea of the method is to build an incremental Delaunay
triangulation use a sample sequence and compute a feedback control in
all collision-free simplices.
The idea can be used for both planning in static environments and
real-time replanning in dynamic environments.

We have established theoretical convergence guarantees and asymptotic
per iteration computational complexity.
%
%
During the experiments in simulated environments, we confirmed
theoretical results and compared the performance of our implementation
with that of state-of-the-art asymptotically optimal planners. It was
established that the performance of a our basic implementation is
similar to that of the previous planning algorithms. However,
optimizing sampling and wavefront propagation routines proved
beneficial for a substantial performance increase.


While being closely related to the previous asymptotically-optimal
graph-based planners, such as, \RRTstar, \RRTsharp, and \RRTx, the
ACIDIC method is fundamentally different from all of them. In
particular, instead of computing an one-dimensional path on a graph,
our method computes a collision-free feedback control that stabilizes
the system towards the goal in the entire volume of the free
space. Thus the output of the ACIDIC method can be used directly to
control the system, and the implementation of auxiliary path-following
controllers can be avoided.
Analysis and experiments show that the ACIDIC method is theoretically
sound and works well in practice.

\addtolength{\textheight}{-4cm}

\section*{ACKNOWLEDGMENT}
We would like to thank Prof. Yuliy Baryshnikov from University of
Illinois at Urbana-Champaign for introducing us to integral geometry,
which enabled asymptotic complexity analysis of the proposed
algorithm.

\bibliographystyle{IEEEtranS}
\bibliography{dmitry_s_yershov}

\end{document}